\documentclass[letterpaper]{article} 
\usepackage{aaai25}  
\usepackage{times}  
\usepackage{helvet}  
\usepackage{courier}  
\usepackage[hyphens]{url}  
\usepackage{graphicx} 
\usepackage{natbib}  
\usepackage{caption} 
\usepackage{algorithm}
\usepackage{algorithmic}

\usepackage{newfloat}
\usepackage{listings}
\DeclareCaptionStyle{ruled}{labelfont=normalfont,labelsep=colon,strut=off} 
\floatstyle{ruled}
\newfloat{listing}{tb}{lst}{}
\floatname{listing}{Listing}
\usepackage{amsmath}
\usepackage{amsfonts}
\usepackage{booktabs}
\usepackage{multirow}
\usepackage{placeins}

\title{KALAHash: Knowledge-Anchored Low-Resource Adaptation for Deep Hashing}
\author{
    Shu Zhao\textsuperscript{\rm 1}, Tan Yu\textsuperscript{\rm 2}, Xiaoshuai Hao\textsuperscript{\rm 3}, Wenchao Ma\textsuperscript{\rm 1}, Vijaykrishnan Narayanan\textsuperscript{\rm 1}
}
\affiliations{
    \textsuperscript{\rm 1}The Pennsylvania State University,\\
    \textsuperscript{\rm 2}NVIDIA,\\
    \textsuperscript{\rm 3}Samsung\\
    \{smz5505, vijaykrishnan.narayanan\}@psu.edu

}

\begin{document}

\maketitle

\begin{abstract}
Deep hashing has been widely used for large-scale approximate nearest neighbor search due to its storage and search efficiency. However, existing deep hashing methods predominantly rely on abundant training data, leaving the more challenging scenario of low-resource adaptation for deep hashing relatively underexplored. This setting involves adapting pre-trained models to downstream tasks with only an extremely small number of training samples available. Our preliminary benchmarks reveal that current methods suffer significant performance degradation due to the distribution shift caused by limited training samples. To address these challenges, we introduce \texttt{C}lass-Calibration \texttt{LoRA}~(CLoRA), a novel plug-and-play approach that dynamically constructs low-rank adaptation matrices by leveraging class-level textual knowledge embeddings. CLoRA effectively incorporates prior class knowledge as anchors, enabling parameter-efficient fine-tuning while maintaining the original data distribution. Furthermore, we propose \texttt{K}nowledge-Gu\texttt{id}ed \texttt{D}iscrete \texttt{O}ptimization~(KIDDO), a framework to utilize class knowledge to compensate for the scarcity of visual information and enhance the discriminability of hash codes. Extensive experiments demonstrate that our proposed method, \texttt{K}nowledge-\texttt{A}nchored \texttt{L}ow-Resource \texttt{A}daptation Hashing~(KALAHash), significantly boosts retrieval performance and achieves a $4\times$ data efficiency in low-resource scenarios.
\end{abstract}

%
\begin{links}
    \link{Code}{https://github.com/Tree-Shu-Zhao/KALAHash.pytorch}
\end{links}

\section{Introduction}

\begin{figure}[t]
	\centering
	\includegraphics[width=1.0\columnwidth]{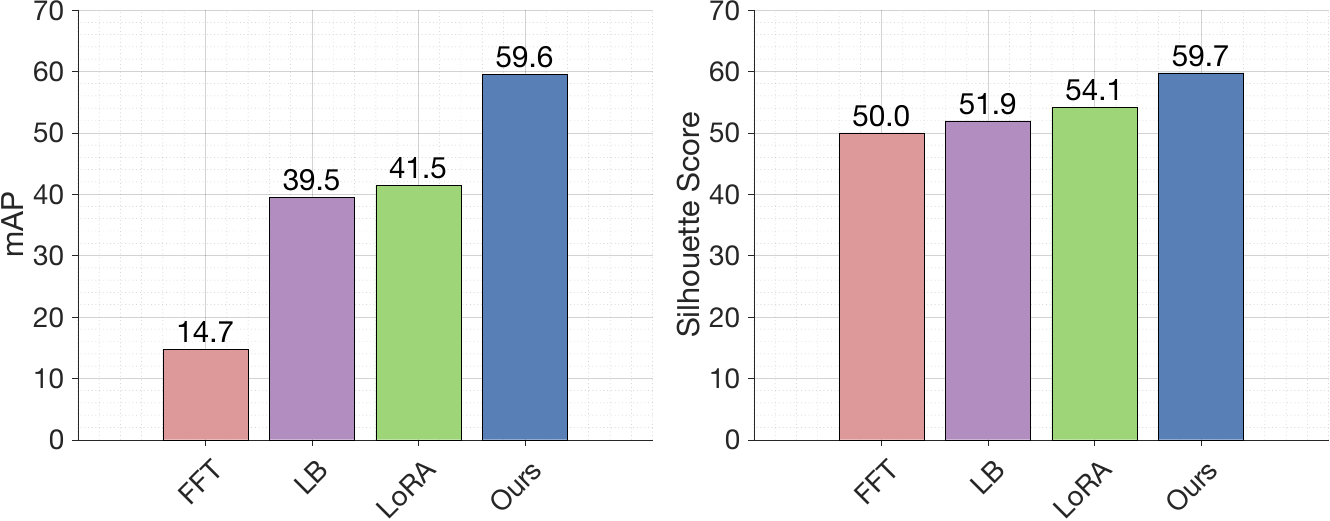}
	\caption{Performance comparison in low-resource settings~(1-shot on the CIFAR-10 dataset), including mean Average Precision scores~(left) and Silhouette Scores~(right). FFT and LB denote Full Fine-Tuning and Lock Backbone, respectively. The increasing mAP and Silhouette Score indicate improved cluster separation and cohesion in the embedding space, demonstrating the effectiveness of our approach in addressing the distribution shift challenge. For the Silhouette Score, we normalize its range from $[-1, +1]$ to $[0, 100]$.}
	\label{fig:degradation}
\end{figure}

Deep hashing has emerged as a powerful technique for large-scale approximate nearest neighbor search, offering significant advantages in terms of storage efficiency and search speed~\cite{HashingSurvey}. While deep hashing methods have shown remarkable performance, they typically rely on the availability of large amounts of data for effective training, which has been a cornerstone of their success but also presents limitations in scenarios where data availability is constrained.

In this paper, we introduce a challenging scenario: low-resource adaptation for deep hashing. This setting is characterized by the need to adapt pre-trained models to the hashing task with extremely limited data samples available for training. The importance of this research direction is twofold. First, it addresses the critical need for efficiency and cost-effectiveness in developing retrieval systems. Annotating large datasets is often prohibitively expensive and time-consuming, especially in specialized domains~\cite{ICCVW_FSH}. By focusing on low-resource adaptation, we aim to reduce the resources required for effective retrieval systems while maintaining high performance. Second, this approach enables rapid adaptation to new domains or emerging topics, a crucial capability in today's fast-paced information landscape~\cite{DBLP:conf/eccv/CohenGMCA22}.

Despite its practical importance, this problem has received relatively little attention in the research community. Our preliminary benchmarks reveal significant challenges in low-resource adaptation for deep hashing. Specifically, we observe substantial performance degradation in existing methods when faced with limited training samples, as illustrated in Figure~\ref{fig:degradation}. Full Fine-Tuning~(FFT) achieves a mean Average Precision~(mAP) of only $14.7\%$. While Lock Backbone~(LB) shows improvements, it also limits the ability of model fine-tuning  and achieves $39.5\%$ mAP. Even equipped with LoRA~\cite{ICLR_LORA}, an advanced technique for enabling parameter-efficient fine-tuning, it still falls short with an mAP score of $41.5\%$. We argue that the performance gap is primarily attributed to the distribution shift, a mismatch between the data distributions of the pre-trained and downstream tasks, occurring when models trained on large datasets are adapted to downstream tasks with scarce data. To measure how this issue affects the distributions of hash codes in hamming space, we employ the Silhouette Score~\cite{rousseeuw1987silhouettes} to measure how similar a class is to its own cluster compared to others. FFT achieves a Silhouette Score of $50.0\%$, denoting the embedding space has collapsed and all data points are close together. While LB and LoRA improve the Silhouette Score, they still cannot perform satisfactorily. The results further underscore the challenge of maintaining cohesive and well-separated clusters in the embedding space under low-resource settings, highlighting the need for more sophisticated adaptation strategies.

Therefore, we recognize the need for a novel approach that can leverage additional sources of information to compensate for the scarcity of data. Recent advancements in Vision-Language Models~(VLMs) have demonstrated their ability to capture rich semantic relationships between visual concepts and textual descriptions~\cite{ICML_CLIP,liu2023llava}. These models, pre-trained on vast amounts of image-text pairs, encapsulate a wealth of class-level knowledge that can potentially guide the adaptation process in low-resource settings. By tapping into this pre-existing knowledge, we hypothesize that we can mitigate the effects of distribution shift and enhance the discriminative power of hash codes, even when faced with limited training samples. 

Motivated by this, we leverage the knowledge within pre-trained VLMs and propose \texttt{C}lass-Calibration \texttt{LoRA}~(CLoRA), a novel plug-and-play approach that dynamically constructs low-rank adaptation matrices by leveraging class-level textual knowledge embeddings. It effectively incorporates prior class knowledge as anchors, enabling parameter-efficient fine-tuning while maintaining the original data distribution. Additionally, we introduce \texttt{K}nowledge-Gu\texttt{id}ed \texttt{D}iscrete \texttt{O}ptimization~(KIDDO), a framework that utilizes class knowledge to compensate for the scarcity of visual information and enhance the discriminability of hash codes.

The main contributions of our work are as follows:
\begin{itemize}
    \item We introduce and benchmark the problem of low-resource adaptation in deep hashing, highlighting its importance and challenges. Our benchmarks reveal significant performance degradation in existing methods when faced with limited training samples.
    \item We propose CLoRA, a novel plug-and-play approach that leverages textual knowledge embeddings as anchors for efficient adaptation in low-resource scenarios.
    \item We develop KIDDO, a knowledge-guided optimization framework that injects knowledge into the optimization process to enhance hash code generation.
    \item We demonstrate that our proposed method significantly improves retrieval performance in challenging low-resource settings through extensive experiments.
\end{itemize}

\section{Related Work}
\noindent\textbf{Deep Hashing for Efficient Retrieval.} Deep hashing has emerged as a powerful approach for large-scale visual retrieval, leveraging deep learning to project high-dimensional data into compact binary codes. The field has evolved from early two-stage methods like CNNH~\cite{AAAI_CNNH} to end-to-end frameworks such as DHHN~\cite{CVPR_DNNH}, which enabled simultaneous optimization of networks and hash codes. The loss functions utilized in deep hashing can be categorized into ranking-based~\cite{ACCV_DTSH,CVPR_TALR}, pair-wise~\cite{IJCAI_DPSH,ICCV_Hashnet,rdh}, and point-wise methods~\cite{CVPR_CSQ,NeurIPS_OrthoHash,CVPR_MDSH}. To address the challenge of discrete optimization, methods like DSDH~\cite{NeurIPS_DSDH} have proposed direct optimization of binary codes using techniques such as discrete cyclic coordinate descent. Architectural innovations, particularly asymmetric designs introduced by DAPH~\cite{MM_DAPH} and further developed in ADSH~\cite{AAAI_ADSH}, CCDH~\cite{MM_CCDH}, and CEDIH~\cite{AAAI_CEDIH}, have significantly improved hash learning quality and efficiency. Despite these advancements, challenges remain in scenarios with limited data. UGH~\cite{ICCVW_FSH} devises a three-phase framework for the few-shot hashing. However, it needs to maintain a large hash function pool and select specific components during inference, which significantly increases the inference delay. Moreover, UGH cannot correctly select components under extremely low-resource adaptation settings, leading to significant performance degradation. Our method leverages the knowledge within the pre-trained models as anchors and complementary information to boost performance under low-resource adaptation settings.

\noindent\textbf{Low-Resource Adaptation.} Low-resource adaptation has gained significant attention in various machine learning domains, addressing scenarios with limited data for multi-modal large language model fine-tuning~\cite{liu2023llava,rebq,neucore,elegant,DBLP:conf/nips/HaoZ23,DBLP:conf/cvpr/HaoZWZL23}. Few-shot learning approaches, such as prototypical networks~\cite{DBLP:conf/nips/SnellSZ17} and MAML~\cite{ICML_MAML}, have pioneered tackling low-resource scenarios by learning transferable knowledge that can quickly adapt to new tasks with minimal data. Recently, low-rank adaptation~(LoRA)~\cite{ICLR_LORA} have demonstrated efficient parameter-tuning for large models. This approach has been particularly effective in natural language processing and is gaining traction in vision tasks. Model merging~\cite{pan2023stitchable,DBLP:journals/corr/abs-2305-02279,DBLP:conf/iclr/YangW00G0T24} combines several model weights trained on different tasks to create a new weight that can perform all tasks simultaneously. Our work focuses on how to train a model for a specific task to achieve better performance. In the specific domain of deep hashing, low-resource adaptation remains relatively unexplored. While methods like \citet{DBLP:conf/cvpr/VenkateswaraECP17} have addressed domain adaptation for hashing, they typically assume a substantial amount of target domain data. The challenge of adapting hash functions with extremely limited data presents a significant research opportunity.

\begin{figure}[t!]
    \centering
    \includegraphics[width=1.0\columnwidth]{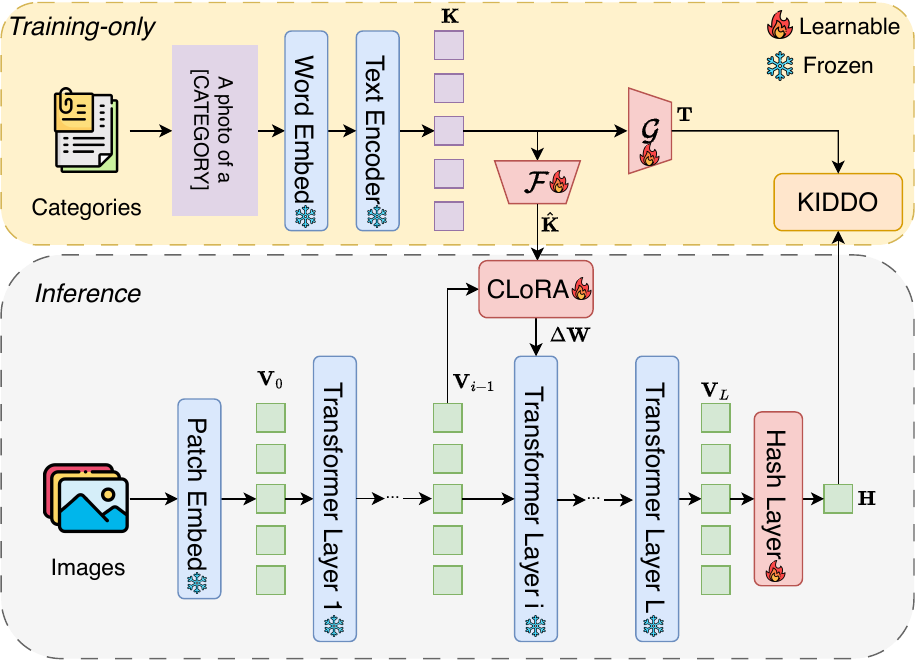}
    \caption{Architecture overview of the proposed KALAHash method, illustrating the integration of Class-Calibration LoRA (CLoRA) and Knowledge-Guided Discrete Optimization (KIDDO).}
    \label{fig:arch}
\end{figure}

\section{Method}

\subsection{Problem Formulation}
Assuming models have been pre-trained on several large source datasets, our goal is to adapt these pre-trained models to learn a hash function that maps images to binary codes while preserving semantic similarity with an extremely small training set $\mathbb{D} = \{\mathbf{x}_i, \mathbf{y}_i\}_{i=1}^{N}$, including a set of $N$ images and their labels.

\subsection{LoRA Background}
LoRA~\cite{ICLR_LORA} is an efficient method for fine-tuning large models. It works by introducing small, trainable matrices into layers of a Transformer model~\cite{NeurIPS_Transformer}. In a standard fully-connected layer, the output is calculated as 
\begin{equation} \label{eq:linear}
        \mathbf{o} = \mathbf{W}\mathbf{x},
\end{equation}
where $\mathbf{W} \in \mathbb{R}^{d\times k}$ is the pre-trained weight matrix; $\mathbf{x} \in \mathbb{R}^{k\times 1}$ is an input vector; $\mathbf{o} \in \mathbb{R}^{d\times 1}$ is the output vector. LoRA modifies Equation~\eqref{eq:linear} by adding a low-rank update: 
\begin{equation} \label{eq:lora}
        \hat{\mathbf{o}} = \mathbf{W}\mathbf{x} + \Delta \mathbf{W}\mathbf{x} = \mathbf{W}\mathbf{x} + \eta\mathbf{P}\mathbf{Q}\mathbf{x}.
\end{equation}
Here, $\mathbf{Q} \in \mathbb{R}^{r \times k}$ and $\mathbf{P} \in \mathbb{R}^{d \times r}$ are small matrices that form the low-rank update. $r \ll \operatorname{min}(k, d)$. $\eta$ is a scale factor. The key is that the number of parameters in $\mathbf{P}$/$ \mathbf{Q}$ is much smaller than the original weight matrix $\mathbf{W}$. During fine-tuning, only $\mathbf{Q}$ and $\mathbf{P}$ are updated, while the original model weights remain frozen. This parameter-efficient approach allows for quick adaptation of large models to new tasks with minimal additional parameters.

However, LoRA, while efficient for parameter updates, does not inherently incorporate task-specific knowledge or constraints. Its generic adaptation mechanism lacks the guidance needed to effectively map high-dimensional image features to compact binary hash codes, especially when provided with only a handful of examples per class, as illustrated in Figure~\ref{fig:degradation}. For deep hashing, especially with limited data, additional guidance about class relationships or desired hash code properties are crucial for generating discriminative hash codes.

To address these limitations and provide the necessary task-specific guidance, we propose leveraging class-level textual knowledge. This approach aims to inject semantic information directly into the adaptation process, bridging the gap between the limited visual data and the rich semantic understanding required for effective hash code generation. By incorporating textual descriptions of image categories, we can provide additional context and structure to guide the learning process, even in extremely low-resource scenarios. This textual knowledge serves as a form of prior information, helping to constrain the adaptation process and ensure that the resulting hash codes maintain semantic relevance. In the following section, we detail our method for extracting and utilizing this class-level textual knowledge to enhance the deep hashing process.

\subsection{Overview}
Figure~\ref{fig:arch} illustrates the architecture of the proposed method. We build our approach on the pre-trained CLIP model~\cite{ICML_CLIP}, including a text and a vision encoder consisting of multiple transformer layers.

The Text Encoder pre-extracts class-level textual knowledge $\mathbf{K}$ using category names. The Vision Encoder splits images into fixed-size patches which are projected into patch embeddings $\mathbf{V}_0$ by the Patch Embed module, and encodes $\mathbf{V}_0$ to vision tokens $\mathbf{V}_L$ by transformer layers, where $L$ denotes the number of transformer layers. During the encoding process, we introduce Class-Calibration LoRA~(CLoRA) module to dynamically construct a weight adjustment matrix $\Delta \mathbf{W}$ by incorporating mapped knowledge $\hat{\mathbf{K}}$ and input vision tokens $\mathbf{V}_{i-1}$ from the $i$-th transformer layer to guide the fine-tuning process. For simplicity, we will omit the subscripts of vision tokens in the following sections. Then, the vision tokens $\mathbf{V}$ are mapped into hash features $\mathbf{H}$. To further improve the hash code generation, we employ Knowledge-Guided Discrete Optimization~(KIDDO), a framework that injects the mapped textual knowledge $\mathbf{T}$ into the optimization process.

\subsection{Class-Level Textual Knowledge Generation}
We use the Text Encoder to pre-extract class-level textual knowledge:
\begin{equation}
\label{keq}
    \mathbf{K} = [\mathbf{k}_1, \mathbf{k}_2, \cdots, \mathbf{k}_C]^{\top} \in \mathbb{R}^{C \times d_t},
\end{equation}
where $C$ is the number of categories. Specifically, we create a prompt based on the hand-crafted template ``\texttt{a photo of a [CATEGORY]}.'' For instance, given a category name \texttt{dog}, the prompt is instantiated as ``\texttt{a photo of a dog}.'' Next, the Word Embed module and Text Encoder map the prompt into a class-level textual knowledge embedding $\mathbf{k}_i$. Note that the konwledge generation only needs to be performed once in the whole process.

\subsection{Class-Calibration LoRA}

We observe that the weight adjustment matrix $\Delta\mathbf{W}$ in Equation~\eqref{eq:lora} can be constructed by:
\begin{equation}
        \Delta\mathbf{W} = \eta\mathbf{P}\mathbf{Q} = \eta\sum_{i=1}^{r}\mathbf{p}_i\mathbf{q}_i^T,
\end{equation}
where $\mathbf{q}_i \in \mathbb{R}^{k \times 1}$, $\mathbf{p}_i \in \mathbb{R}^{d \times 1}$.

\begin{figure}[t]
    \centering
    \includegraphics[width=0.4\textwidth]{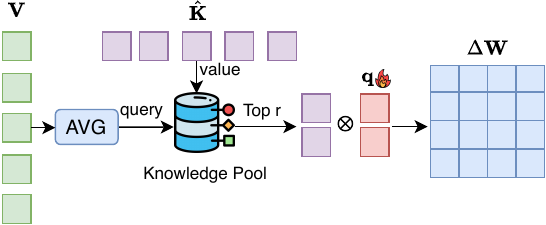}
    \caption{Architecture of the proposed CLoRA module.}
    \label{fig:clora}
\end{figure}

As shown in Figure~\ref{fig:clora}, to constrain the weight adjustment matrix space, spanning by $\mathbf{p}_i\mathbf{q}_i^T$, we replace $\mathbf{p}_i$ with the class-level textual knowledge $\mathbf{k}_i$ defined in Equation ~(\ref{keq}) as anchors:

\begin{equation}
        \Delta\mathbf{W} = \eta\sum_{i=1}^{r}\hat{\mathbf{k}}_i\mathbf{q}_i^T,
\end{equation}
where $\hat{\mathbf{k}}_i = \mathcal{F}(\mathbf{k}_i)$, $\mathcal{F}$ is a linear layer and $\mathcal{F}(\cdot) \in \mathbb{R}^{d\times 1}$. 

Considering that different inputs need different knowledge, we design a query-based strategy to dynamically select $r$ knowledge vectors from the knowledge pool $\hat{\mathbf{K}}$:

\begin{equation} \label{eq:select}
        \hat{\mathbf{K}}^{v} = \operatorname{Top}_r(\operatorname{avg}(\mathbf{V}), \hat{\mathbf{K}}),
\end{equation}
where $\mathbf{V}=[\mathbf{v}_1,\cdots, \mathbf{v}_t]$ is the vision tokens; $\operatorname{avg}(\cdot)$ denotes the average pooling operation. $\operatorname{Top}_r(\cdot,\cdot)$ selects the top $r$ vectors in $\hat{\mathbf{K}}$ with the largest cosine similarity to $\operatorname{avg}(\mathbf{V})$, and $\hat{\mathbf{K}}^{v} = [\hat{\mathbf{k}}^{v}_1, \cdots, \hat{\mathbf{k}}^{v}_r]$.

Finally, the weight adjustment matrix is constructed by:
\begin{equation} \label{eq:weight_adjustment}
        \Delta\mathbf{W} = \eta\sum_{i=1}^{r}\hat{\mathbf{k}}^{v}_i\mathbf{q}_i^T.
\end{equation}

\subsection{Knowledge-Guided Discrete Optimization}

We first employ a similarity loss function $\mathcal{L}_{\textrm{s}}$ and a quantization loss $\mathcal{L}_{\textrm{q}}$ that are widely used in deep hashing methods, which makes the Hamming distance of two similar points as small as possible and vice versa:

\begin{equation} \label{eq:sim_qua_loss}
    \begin{aligned}
        \mathcal{L}_{\textrm{s}} &= -\sum_{s_{ij} \in \mathbf{S}}\left(s_{ij}\theta_{ij} - \log\left(1 + e^{\theta_{ij}}\right)\right),\\
        \mathcal{L}_{\textrm{q}} &= \Vert\mathbf{H}-\mathbf{B}\Vert_2^2,
    \end{aligned}
\end{equation}
where $s_{ij}$ is $1$ if image $i$ and $j$ belong to the same category otherwise $0$; $\theta_{ij} = \frac{1}{2}\mathbf{h}^{\top}_i\mathbf{h}_j$; $\mathbf{H} = [\mathbf{h}_1,  \cdots, \mathbf{h}_n]^{\top}$ are real-value image features generated from Hashing layer; $\mathbf{B} = [\mathbf{b}_1,  \cdots, \mathbf{b}_n]^{\top}$, $\mathbf{b}_i \in \{-1, 1\}^b$ is the learned binary code and is randomly initialized before the training.

In low-resource settings, the limited number of training images may not be sufficient to cover all aspects of visual concepts, thus leading to over-fitting issues. We argue that language can be used as an abstract conceptual representation as anchor points to aid visual feature learning. For instance, while ``dog'' is visually represented in various ways that cannot all be covered by a limited number of images, they can be abstracted into a single linguistic concept ``dog''.

Motivated by this, we add an alignment loss $\mathcal{L}_a$ between the learned binary codes $\mathbf{B}$ and the textual knowledge $\mathbf{K}$ to further improve the hash code generation by leveraging the textual knowledge as anchors:

\begin{equation} \label{eq:align_loss}
        \mathcal{L}_{\textrm{a}} = \Vert\mathbf{Y}- \mathbf{T}^{\top}\mathbf{B}\Vert_2^2,
\end{equation}
where $\mathbf{T}= \mathcal{G}(\mathbf{K})$, $\mathcal{G}(\cdot)$ denotes a fully-connected layer and $\mathbf{Y} = [\mathbf{y}_1, \cdots, \mathbf{y}_n] \in \mathbb{R}^{C \times n}$ are one-hot label vectors where $C$ is the number of categories.

Finally, the loss function is
\begin{equation} \label{eq:loss}
        \mathcal{L} = \alpha\mathcal{L}_{\textrm{a}} + \beta\mathcal{L}_{\textrm{q}} + \gamma\mathcal{L}_{\textrm{s}},
\end{equation}
where $\alpha$, $\beta$, and $\gamma$ are scalars that balance the loss values.

When optimizing the loss function $\mathcal{L}$ in Equation~\eqref{eq:loss}, it not only exploits the knowledge from the text encoder as complementary information to improve the image hash code generation but also enables the possibility of discrete optimization. To optimize the loss function in Equation~\eqref{eq:loss}, we use the standard backpropagation algorithm to learn $\mathbf{H}$ and $\mathbf{T}$. To optimize $\mathbf{B}$, we fix all the variables except for $\mathbf{B}$ and rewrite the optimization formula as
\begin{equation} \label{eq:dcc}
    \begin{aligned}
        &\min_{\mathbf{B}} \alpha\left\Vert \mathbf{Y} - \mathbf{T}^{\top}\mathbf{B} \right\Vert^2_2 + \beta \left\Vert \mathbf{H} - \mathbf{B} \right\Vert^2_2\\
		&\qquad\text{s.t. } \mathbf{B} \in \{-1, 1\}^{N\times b}.
    \end{aligned}
\end{equation}
Then, we adopt the discrete cyclic coordinate descent~(DCC) method proposed by \citet{CVPR_SDH} to optimize $\mathbf{B}$ column by column. The optimal solution of Equation~\eqref{eq:dcc} is
\begin{equation} \label{eq:dcc_solution}
        \mathbf{B}^{i} = \operatorname{sign}(\mathbf{S}^i - \mathbf{B}^{'\top}\mathbf{T}^{'}\mathbf{T}^i),
\end{equation}
where $\mathbf{B}^i$ is the $i^{\text{th}}$ column of $\mathbf{B}$, $\mathbf{B}^{'}$ is the matrix of $\mathbf{B}$ excluding $\mathbf{B}^i$; $\mathbf{S}^i$ is the $i^{\text{th}}$ row of matrix $\mathbf{S}$, $\mathbf{S} = \beta\mathbf{Y}\mathbf{T} + \gamma\mathbf{H}$, $\mathbf{S}^{'}$ is the matrix of $\mathbf{S}$ excluding $\mathbf{S}^i$; $\mathbf{T}^i$ is the $i^{\text{th}}$ row of $\mathbf{T}$, $\mathbf{T}^{'}$ is the matrix of $\mathbf{T}$ excluding $\mathbf{T}^i$.

In Equation~\eqref{eq:dcc_solution}, textual knowledge is injected into binary code $\mathbf{B}$, further improving the optimization process of hash code generation $\mathbf{H}$. In the following section, we will demonstrate the effectiveness of our proposed method.

\begin{table*}[ht]
	\centering
	\resizebox{1\textwidth}{!}{
		\begin{tabular}{l|cccc|cccc|cccc}
			\toprule
			\multirow{2}{*}{Method} & \multicolumn{4}{c}{NUS-WIDE} & \multicolumn{4}{c}{MS-COCO} & \multicolumn{4}{c}{CIFAR-10} \\
			& 1-shot & 2-shot & 4-shot & 8-shot & 1-shot & 2-shot & 4-shot & 8-shot & 1-shot & 2-shot & 4-shot & 8-shot\\
                \midrule
			HashNet~\cite{ICCV_Hashnet} & 65.23 & 66.27 & 70.54 & 73.56 & 58.81 & 62.44 & 65.28 & 67.50 & 41.68 & 44.97 & 69.96 & 76.58\\
			DSDH~\cite{NeurIPS_DSDH} & 67.32 & 69.23 & 72.15 & 74.13 & 59.63 & 62.44 & 66.84 & 68.72 & 44.22 & 53.14 & 71.76 & 77.26\\
                DCH~\cite{CVPR_DCH} & 65.55 & 66.04 & 70.92 & 71.48 & 60.32 & 62.26 & 66.51 & 67.70 & 39.53 & 48.69 & 67.03 & 75.59\\
			GreedyHash~\cite{NeurIPS_GreedyHash} & 67.24 & 69.96 & 71.71 & 72.21 & 59.81 & 63.91 & 65.84 & 70.28 & 44.87 & 57.01 & 72.00 & 77.58\\
                CSQ~\cite{CVPR_CSQ} & 65.75 & 67.31 & 70.96 & 71.51 & 59.23 & 63.09 & 66.14 & 70.18 & 46.66 & 60.54 & 69.50 & 77.69\\
                OrthoHash~\cite{NeurIPS_OrthoHash} & 67.31 & 70.96 & 71.48 & 71.59 & 60.21 & 64.13 & 66.34 & 70.23 & 46.68 & 60.03 & 73.37 & 77.63\\ 
                HSWD$^{\dag}$~\cite{CVPR_SWD} & 67.58 & 67.83 & 70.44 & 74.10 & 60.15 & 62.86 & 66.28 & 69.06 & 48.63 & 57.36 & 73.24 & 79.37\\
                MDSH$^{\ddag}$~\cite{CVPR_MDSH} & 67.23 & 68.22 & 70.47 & 72.04 & 58.55 & 59.89 & 60.94 & 63.95 & 47.33 & 58.69 & 73.16 & 78.09\\ 
                \midrule
			KALAHash& \textbf{70.69} & \textbf{71.26} & \textbf{74.11} & \textbf{75.24} & \textbf{65.32} & \textbf{66.43} & \textbf{71.98} & \textbf{73.96} & \textbf{57.54} & \textbf{70.00} & \textbf{80.14} & \textbf{83.00}\\
			\bottomrule
	\end{tabular}}
	\caption{Comparison of mAP on NUS-WIDE, MS-COCO, and CIFAR-10 datasets for different deep hashing methods under various low-resource settings~(1-shot to 8-shot).$\dag$: we use the HashNet-HSWD. $\ddag$: MSDH conducted experiments only on single-label datasets in the original paper.}
	\label{tab:main_results}
\end{table*}

\begin{table}[t]
\centering
\resizebox{0.8\linewidth}{!}{
\begin{tabular}{@{}l|ccc@{}}
\toprule
Method & NUS-WIDE & MS-COCO & CIFAR-10 \\ 
\midrule
HashNet & 65.23 & 58.81 & 41.68\\
+CLoRA & \textbf{69.41} & \textbf{61.08} & \textbf{54.20} \\ 
\midrule
DSDH & 67.32 & 59.63 & 44.22 \\
+CLoRA & \textbf{70.02} & \textbf{62.43} & \textbf{54.02} \\ 
\midrule
DCH & 65.55 & 60.32 & 39.53 \\
+CLoRA & \textbf{69.48} & \textbf{61.83} & \textbf{50.55} \\ 
\midrule
GreedyHash & 67.24 & 59.81 & 44.87 \\
+CLoRA & \textbf{70.30} & \textbf{60.74} & \textbf{51.77} \\ 
\midrule
CSQ & 65.75 & 59.23 & 46.66 \\
+CLoRA & \textbf{69.14} & \textbf{60.54} & \textbf{49.44} \\ 
\midrule
OrthoHash & 67.31 & 60.21 & 49.50 \\
+CLoRA & \textbf{69.61} & \textbf{61.39} & \textbf{51.58} \\ 
\midrule
HSWD & 67.58 & 58.55 & 48.63\\
+CLoRA & \textbf{68.85} & \textbf{60.75} & \textbf{52.63} \\
\midrule
MDSH & 67.23 & 58.55 & 47.33 \\
+CLoRA & \textbf{68.24} & \textbf{60.34} & \textbf{48.29} \\ 
\bottomrule
\end{tabular}
}
\caption{Plug-and-play capability of CLoRA. mAP improvements when applying CLoRA to various baseline deep hashing methods on NUS-WIDE, MS-COCO, and CIFAR-10 datasets.}
\label{tab:plug_and_play}
\end{table}

\section{Experiments}

\subsection{Datasets}
We evaluate our proposed method on three standard benchmarks: NUS-WIDE~\cite{NUS-WIDE}, MS-COCO~\cite{MSCOCO}, and CIFAR-10~\cite{CIFAR10}.

\noindent  \textbf{NUS-WIDE} is a multi-label dataset. Following~\citet{NeurIPS_OrthoHash}, we adopt a subset of the original NUS-WIDE dataset, which has $195,834$ images associated with the $21$ most frequent classes. We randomly select $2,100$ images~($1,00$ images per class) to form the query set, and the rest is used as the gallery set.

\noindent  \textbf{MS-COCO} is a multi-label dataset containing $82,783$ training images and $40,504$ validation images belonging to 80 classes. We combine the two sets of images and prune them without labels. Following~\citet{NeurIPS_OrthoHash}, we randomly choose $5,000$ images as the query set, and the rest are viewed as the gallery set.

\noindent \textbf{CIFAR-10} consists of $60,000$ images with $32 \times 32$ resolution. It has $10$ classes, each containing $6,000$ samples. Following~\citet{CVPR_DCH}, we randomly sample $1,000$ images~($100$ images per class) to construct the query set and the rest is used to form the gallery set.

\subsection{Evaluation Protocol} 
In our low-rank adaptation setting, we randomly split $N_K$ samples per class to create the training set. $N_K$ is $1$, $2$, $4$, or $8$ in our experiments. Following the standard evaluation protocol, we report the mean Average Precision at $K$~(mAP@$K$), the mean of average precision scores of the top $K$ retrieved images, to evaluate the retrieval performance. Specifically, we report mAP@$59000$ for CIFAR-10, mAP@$5000$ for NUS-WIDE, and mAP@$5000$ for MS-COCO, respectively. Notably, for multi-label datasets, two images are considered similar if they share at least one common label.

\subsection{Implementation Details}
For a fair comparison, all methods, including baselines, use the same backbone model, optimizer, training hyper-parameters, etc.

\noindent\textbf{Backbone and CLoRA}. We employ the CLIP ViT-B/32 as the backbone model to conduct experiments. CLoRA can be inserted into various positions in backbones. In our experiments, it is put into the key and value matrices of the multi-head attention module in the last transformer layer. $\eta$ and $r$ are set to $1.0$ and $1$, respectively.

\noindent\textbf{Hash Layer}. A hash layer is utilized to map the original features extracted from the backbone model to compacted hash codes. Following \citet{NeurIPS_OrthoHash}, the hash layer consists of a full-connected layer, a batch norm layer and a $\operatorname{tanh}$ layer. In our experiments, we use a $16$-bit hash layer as default. 

\noindent\textbf{Training Details}. We freeze all the parameters expected for the CLoRA module, $\mathcal{G}$, $\mathcal{F}$, and hash layer. We use SGD with $0.9$ momentum and $1e-5$ weight decay as the optimizer. The learning rate is set to $0.01$. The batch size is set to $8$. $\alpha$, $\beta$, and $\gamma$ are set to $0.1$, $1.0$, and $3.0$, respectively.

If not specified, experiments are conducted on the CIFAR-10 dataset with the 1-shot setting. Detailed settings can be found in the code we provided.

\begin{table}[t]
\centering
\resizebox{0.8\linewidth}{!}{
    \begin{tabular}{r|ccc}
        \toprule
        Method & NUS-WIDE & MS-COCO & CIFAR-10\\
        \midrule
        KALAHash & \textbf{70.69} & \textbf{65.32} & \textbf{57.54}\\
        \midrule
        \textit{w.o.} CLoRA & 68.31 & 61.49 & 46.38\\
        \textit{w.o.} KIDDO & 66.97 & 60.48 & 50.89\\
        \bottomrule
    \end{tabular}
    }
    \caption{Ablation study showing the impact of CLoRA and KIDDO components on mAP performance across NUS-WIDE, MS-COCO, and CIFAR-10 datasets.}
    \label{tab:ablation}
\end{table}
\begin{figure}[t]
    \centering
    \includegraphics[width=0.76\linewidth]{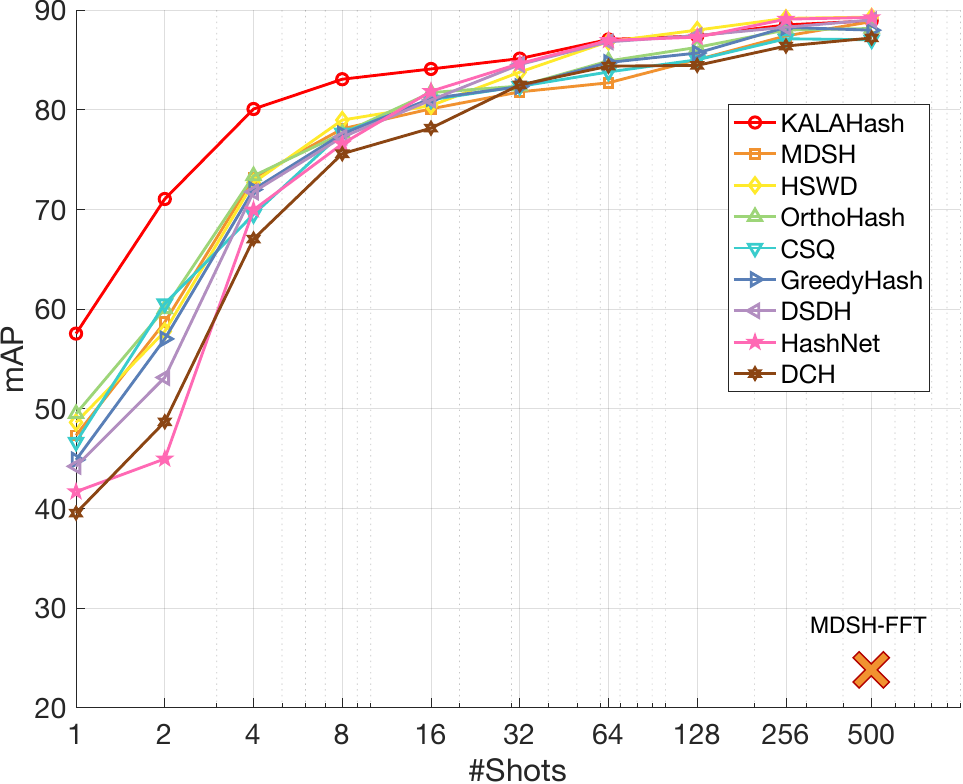}
    \caption{Performance comparison of KALAHash and baseline methods as the number of shots increases from $1$ to $500$ on CIFAR-10 dataset. MDSH-FFT denotes all the parameters are fine-tuned in the MDSH baseline.}
    \label{fig:shots}
\end{figure}
\begin{table}[t]
\centering
\resizebox{1\linewidth}{!}{
    \begin{tabular}{l|ccc}
        \toprule
        Variant & NUS-WIDE & MS-COCO & CIFAR-10\\
        \midrule
        CLoRA & \textbf{70.69} & \textbf{65.32} & \textbf{57.54} \\
        \midrule
        LoRA~\cite{ICLR_LORA} & 68.24 & 61.63 & 47.05 \\
        Prompt Tuning~\cite{IJCV_COOP} & 69.65 & 61.37 & 49.70\\
        Bias Tuning~\cite{ACL_BITFIT} & 69.90 & 63.26 & 45.81\\
        \bottomrule
    \end{tabular}
    }
    \caption{Comparison of CLoRA with other parameter-efficient fine-tuning techniques on NUS-WIDE, MS-COCO and CIFAR-10 datasets.}
    \label{tab:peft}
\end{table}
\begin{figure}[t]
    \centering
    \includegraphics[width=0.8\linewidth]{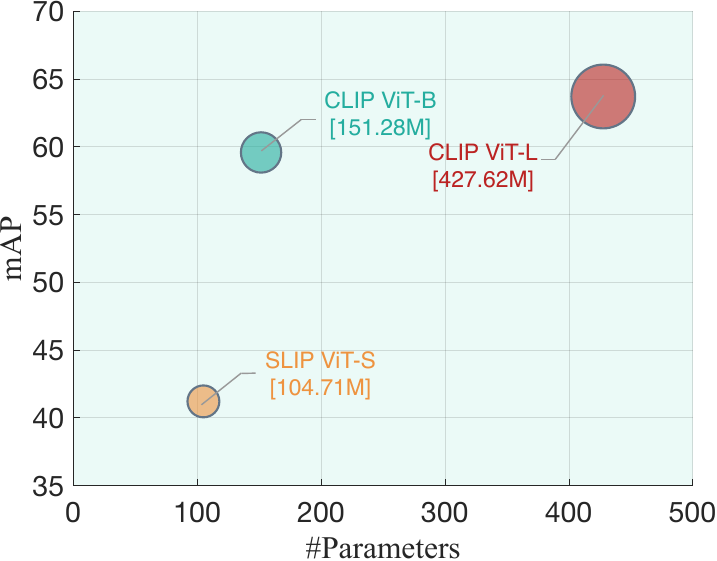}
    \caption{Performance scaling of KALAHash with different backbone models (SLIP ViT-S, CLIP ViT-B, CLIP ViT-L) in relation to the number of model parameters.}
    \label{fig:backbones}
\end{figure}

\subsection{Main Results}

We choose several representative deep hashing methods as baselines, including HashNet~\cite{ICCV_Hashnet}, DSDH~\cite{NeurIPS_DSDH}, DCH~\cite{CVPR_DCH}, GreedyHash~\cite{NeurIPS_GreedyHash}, CSQ~\cite{CVPR_CSQ}, OrthoHash~\cite{NeurIPS_OrthoHash}, HSWD~\cite{CVPR_SWD}, and MDSH~\cite{CVPR_MDSH}.

Table~\ref{tab:main_results} presents the mAP results for NUS-WIDE, MS-COCO, and CIFAR-10 across different low-resource settings~($1$-$8$ shots). We note that SOTA methods do not show absolute competitiveness in low-resource settings as on full-size datasets, highlighting the need for more sophisticated adaptation strategies. Our proposed KALAHash consistently outperforms all baselines across all datasets and shot settings. The performance improvements are particularly significant in the extreme low-resource scenarios ($1$-shot and $2$-shot).
For CIFAR-10, KALAHash achieves $8.91\%$-$18.01\%$ improvements over the baselines in the $1$-shot setting. This performance gap remains substantial even as the number of shots increases, with KALAHash maintaining $3.63\%$-$7.41\%$ improvements in the $8$-shot setting. We can also observe the same trend on the NUS-WIDE and MS-COCO datasets. The multi-label nature of these two datasets highlights KALAHash's ability to handle complex semantic relationships even with limited data.

\subsection{Plug-and-Play Capability}
Table~\ref{tab:plug_and_play} demonstrates KALAHash's plug-and-play capability by applying CLoRA to various baseline methods. Across all baselines and datasets, adding CLoRA consistently improves performance. Specifically, the ranges of improvements are from $1.01\%$-$4.18\%$ on NUS-WIDE, $0.93\%$-$2.80\%$ on MS-COCO, and $0.96\%$-$12.52\%$ on CIFAR-10, respectively. The results underscore the versatility and effectiveness of our proposed method in enhancing existing deep hashing approaches in low-resource scenarios.

\begin{figure*}[t]
    \centering
    \includegraphics[width=0.93\linewidth]{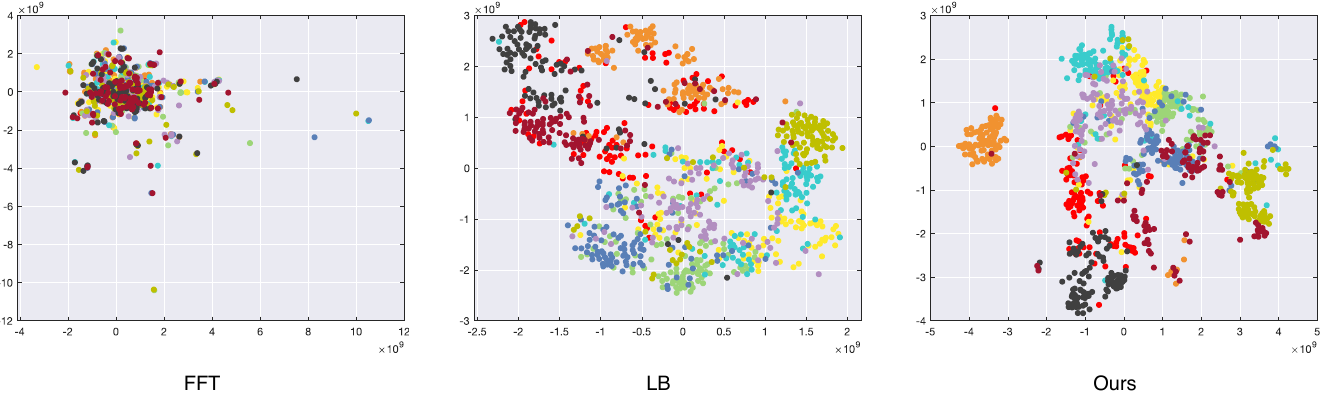}
    \caption{t-SNE visualization of learned hash codes for Full Fine-Tuning (FFT), Lock Backbone (LB), and our proposed method (KALAHash). Different colors denote different categories.}
    \label{fig:tsne}
\end{figure*}
\begin{figure*}[t]
    \centering
    \includegraphics[width=0.95\linewidth]{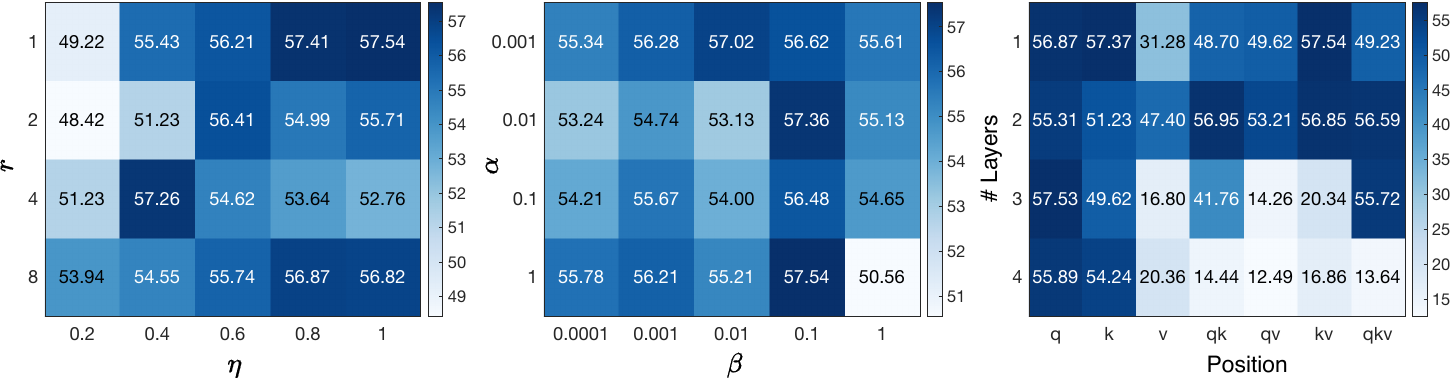}
    \caption{Parameter sensitivity analysis for KALAHash, showing mAP performance across different hyper-parameter settings.}
    \label{fig:sensitivity}
\end{figure*}

\subsection{Ablation Studies}
To understand the contribution of each component in KALAHash, we conduct ablation studies, as shown in Table~\ref{tab:ablation}. Removing CLoRA results in a significant performance drop across all datasets, with mAP decreasing by $2.38\%$, $3.83\%$, and $11.16\%$ on NUS-WIDE, MS-COCO, and CIFAR-10 respectively. Similarly, removing KIDDO leads to performance degradation, with mAP decreasing by $3.72\%$, $4.84\%$, and $6.65\%$ on NUS-WIDE, MS-COCO, and CIFAR-10 respectively, demonstrating the effectiveness of injecting textual knowledge into the optimization process.

We also compare CLoRA to other parameter-efficient fine-tuning techniques in Table~\ref{tab:peft}. CLoRA outperforms standard LoRA~\cite{ICLR_LORA}, Prompt Tuning~\cite{IJCV_COOP}, and Bias Tuning~\cite{ACL_BITFIT} across all datasets, demonstrating its effectiveness.

\subsection{Scalability}

Figure~\ref{fig:shots} presents a comprehensive analysis of KALAHash's performance as the number of shots increases from $1$ to $500$ on CIFAR-10. KALAHash consistently outperforms all baselines with limited training samples~($1$-$16$ shots). As the number of training samples increases, our approach still maintains a performance comparable to SOTA's. This highlights the method's effectiveness in extremely low-resource scenarios while also demonstrating its ability to maintain superior performance as more data becomes available. We note that the SOTA methods do not show absolute competitiveness. The reason may be that we lock the backbone and only fine-tune the FC layer, limiting its ability. However, even full fine-tuning of VLMs on full datasets can also lead to serious distribution shift issue. To demonstrate this, we report the result of MDSH-FFT on the full-size CIFAR-10 dataset as an orange cross, illustrated in Figure~\ref{fig:shots}.

Figure~\ref{fig:backbones} illustrates the performance of KALAHash across different backbone models, including SLIP ViT-S~\cite{ECCV_SLIP}, CLIP ViT-B, and CLIP ViT-L~\cite{ICML_CLIP}. The results show that KALAHash's performance scales well with larger backbone models, achieving higher mAP scores as the number of parameters increases. This demonstrates the method's ability to leverage more powerful pre-trained models effectively.

\subsection{Qualitative Analysis}
Figure~\ref{fig:tsne} provides a t-SNE visualization~\cite{TSNE} of the learned hash codes for Full Fine-Tuning~(FFT), Lock Backbone~(LB), and our proposed method. The results show that the embedding space of FFT is collapsed due to the distribution shift issue. While LB improves this, it still cannot perform satisfactorily, as the red points are scattered in the embedding space. The visualization of KALAHash clearly shows that it produces more compact and well-separated clusters than the other methods. This qualitative result supports our quantitative findings and illustrates KALAHash's ability to learn more discriminative hash codes even in low-resource settings.

\begin{table}[t]
\centering
\resizebox{1.0\columnwidth}{!}{
    \begin{tabular}{@{}rcccc@{}}
        \toprule
        & ViT-S/16 & ViT-B/32 & ViT-B/16 & ViT-L/14 \\ \midrule
        \textit{w.o.} CLoRA & 2.20 $\pm$ 0.03 & 1.17 $\pm$ 0.02 & 2.20 $\pm$ 0.04 & 6.28 $\pm$ 0.04\\
        \textit{w.} CLoRA & 2.21 $\pm$ 0.05 & 1.17 $\pm$ 0.05 & 2.23 $\pm$ 0.02 & 6.33 $\pm$ 0.03\\ \bottomrule
    \end{tabular}
    }
    \caption{Inference time~(ms) per image comparison of various VLMs with and without CLoRA. CLoRA demonstrates negligible impact on inference speeds across different model architectures.}
    \label{tab:inference_time}
\end{table}

\subsection{Parameter Sensitivity}
Figure~\ref{fig:sensitivity} examines the sensitivity of KALAHash to its key hyper-parameters, where ``\#Layers'' denotes the number of layers inserted by CLoRA, and ``Position'' means which attention matrices are adjusted by CLoRA. The results show that KALAHash is robust to parameter changes, maintaining strong performance across a wide range of values.

\subsection{Inference Time}
We conducted a comprehensive analysis of inference times to evaluate the computational efficiency of our proposed CLoRA method across various backbones with and without CLoRA. As shown in Table~\ref{tab:inference_time}, the integration of CLoRA introduces minimal computational overhead across all tested architectures demonstrating that CLoRA maintains the efficiency of the original models while providing the benefits of knowledge-anchored adaptation.

\section{Acknowledgment}
This work was supported in part by Semiconductor Research Corporation JUMP 2.0 PRISM Center.

\bibliography{aaai25}

\clearpage
\FloatBarrier
\appendix
\section*{Appendix}

\begin{figure}[ht]
	\centering
	\includegraphics[width=0.85\columnwidth]{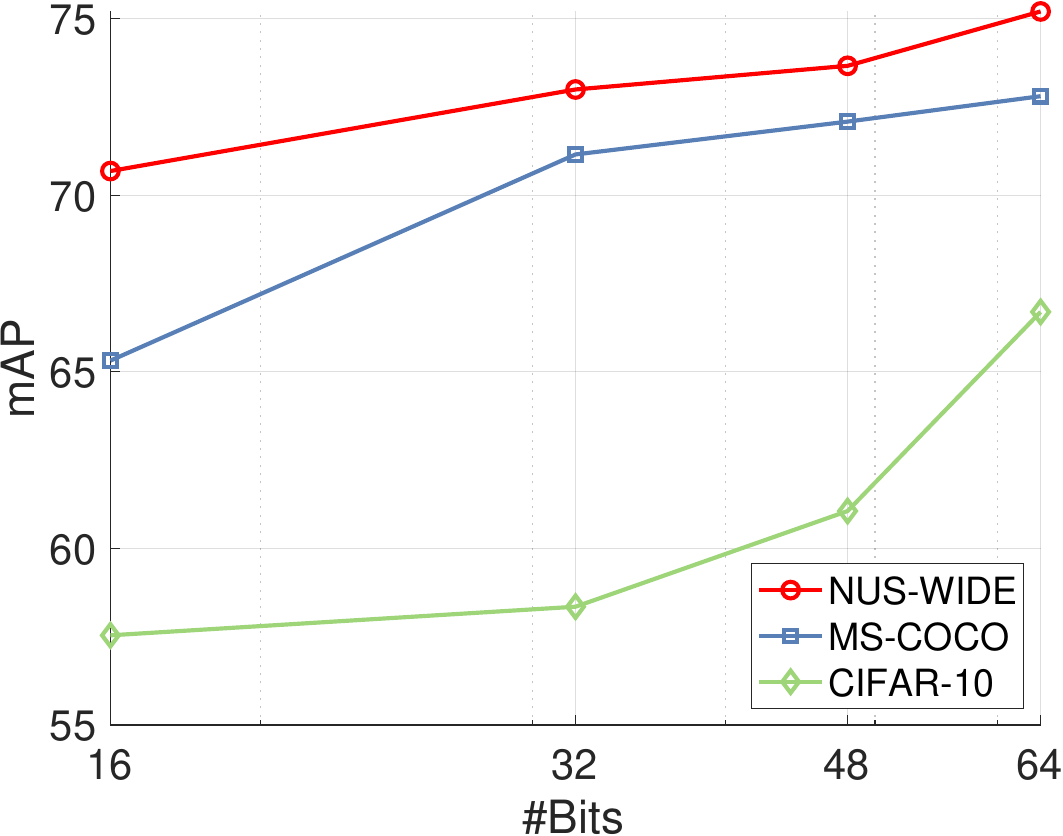}
	\caption{Performance of KALAHash as the number of bits increases from $16$ to $64$ on NUS-WIDE, MS-COCO, and CIFAR-10 datasets.}
	\label{fig:bits}
\end{figure}
\section{Scalability of the Number of Bits}
Figure~\ref{fig:bits} presents a comprehensive analysis of KALAHash's performance as the number of bits increases from $16$ to $64$ on NUS-WIDE, MS-COCO, and CIFAR-10. As the number of training samples increases, our approach consistently improves the retrieval performance, which demonstrates its ability to scale up the number of bits.

\begin{table}[t]
\centering
\resizebox{0.8\linewidth}{!}{
\begin{tabular}{@{}l|c@{}}
\toprule
Method & Silhouette Score \\ 
\midrule
HashNet~\cite{ICCV_Hashnet} & 57.48 \\
DSDH~\cite{NeurIPS_DSDH} & 57.50\\
DCH~\cite{CVPR_DCH} & 55.96\\
GreedyHash~\cite{NeurIPS_GreedyHash} & 52.25\\
CSQ~\cite{CVPR_CSQ} & 52.36\\
OrthoHash~\cite{NeurIPS_OrthoHash} & 51.84\\
HSWD~\cite{CVPR_SWD} & 55.54\\
MDSH~\cite{CVPR_MDSH} & 52.07 \\
\midrule
KALAHash & \textbf{59.76}\\
\bottomrule
\end{tabular}
}
\caption{Silhouette Scores for various hashing methods on CIFAR-10.}
\label{tab:silhouette_score}
\end{table}

\begin{figure}[t]
	\centering
	\includegraphics[width=0.70\columnwidth]{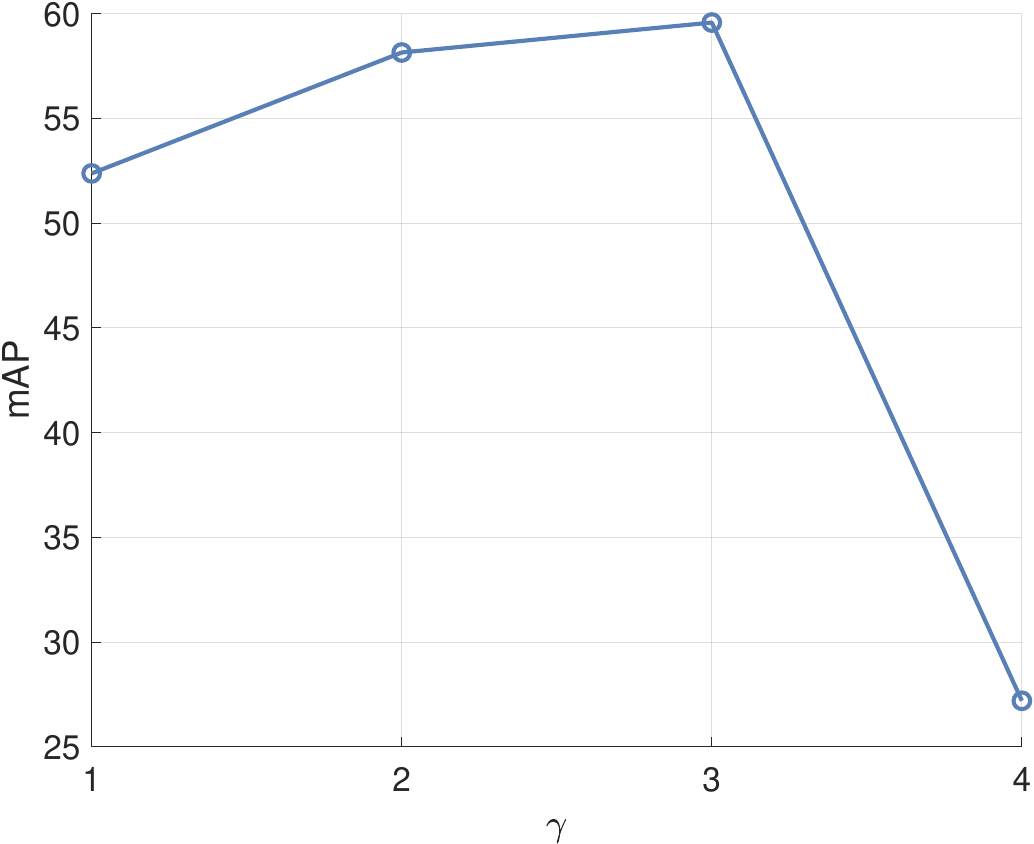}
	\caption{Sensitivity analysis of KALAHash with respect to $\gamma$ on CIFAR-10.}
	\label{fig:gamma}
\end{figure}

\section{Silhouette Score}
Table~\ref{tab:silhouette_score} shows the results of the Silhouette Score. KALAHash consistently outperforms baseline methods, demonstrating the effectiveness of our proposed method. Besides, we notice that methods using pair-wise loss achieve a higher score than those using point-wise loss. The reason may be that contrastive loss has a stronger ability to push hash codes belonging to different classes to different locations in the embedding space. 

\section{Parameter Sensitivity}
Figure~\ref{fig:gamma} examines the sensitivity of KALAHash to $\gamma$. KALAHash maintains relatively high mAP scores when $\gamma \leq 3$. There is a noticeable drop in performance when $\gamma \geq 4$, indicating that extremely high values may affect the optimization progress, leading to a suboptimal result. 

\begin{table*}[t]
\centering
\resizebox{0.77\linewidth}{!}{

\begin{tabular}{@{}l|cccc@{}}
\toprule
Method & ResNet-18 & ResNet-50 & ImageNet21k ViT-B/32 & CLIP ViT-B/32\\ 
\midrule
HashNet~\cite{ICCV_Hashnet} & 15.25 & 23.58 & 31.35 & \textbf{41.68}\\
DSDH~\cite{NeurIPS_DSDH} & 14.27 & 16.24 & 24.50 & \textbf{44.22}\\
DCH~\cite{CVPR_DCH} & 16.17 & 19.08 & 21.79 & \textbf{39.53} \\
GreedyHash~\cite{NeurIPS_GreedyHash} & 16.01 & 19.26 & 24.24 & \textbf{44.87}\\
CSQ~\cite{CVPR_CSQ} & 15.01 & 18.10 & 22.59 & \textbf{46.66}\\
OrthoHash~\cite{NeurIPS_OrthoHash} & 16.01 & 19.73 & 27.56 & \textbf{46.68} \\
HSWD~\cite{CVPR_SWD} & 14.86 & 23.20 & 30.70 & \textbf{48.63}\\
MDSH~\cite{CVPR_MDSH} & 16.19 & 18.08 & 24.28 & \textbf{47.33} \\
\bottomrule
\end{tabular}
}
\caption{Comparison of different hashing methods using various backbone architectures.}
\label{tab:backbones}
\end{table*}

\begin{figure*}[t]
    \centering
    \includegraphics[width=0.83\linewidth]{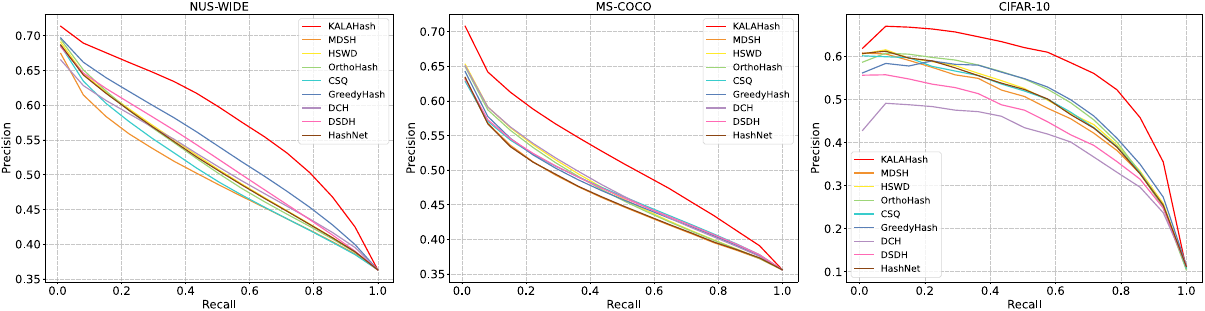}
    \caption{Precision-Recall curves for KALAHash and baseline methods on NUS-WIDE, MS-COCO, and CIFAR-10 datasets.}
    \label{fig:pr_curve}
\end{figure*}

\section{PR Curve}
Figure~\ref{fig:pr_curve} illustrates the Precision-Recall (PR) curves for KALAHash and baseline methods on NUS-WIDE, MS-COCO, and CIFAR-10 datasets. These curves provide a comprehensive view of the model's performance across different precision and recall thresholds. The results demonstrate that KALAHash consistently exhibits competitive performance across all three datasets, maintaining a good balance between precision and recall. The consistent performance across varied datasets highlights the versatility and robustness of our proposed method.

\section{Various Backbones}
Pretrained Backbones are vital for downstream tasks~\cite{DBLP:conf/cvpr/NgZS022,DBLP:conf/eccv/HaoLZLYJPYZZ24,DBLP:journals/corr/abs-2406-12214}. Table~\ref{tab:backbones} presents the performance of various baseline methods across different backbone architectures. We evaluate the methods using ResNet-18~\cite{CVPR_RESNET}, ResNet-50~\cite{CVPR_RESNET}, ImageNet21k ViT-B/32~\cite{ICLR_VIT}, and CLIP ViT-B/32~\cite{ICML_CLIP} as backbone networks. Notably, the performance improves when moving from CNN-based architectures (ResNet-18 and ResNet-50) to transformer-based architectures (ImageNet21k ViT-B/32 and CLIP ViT-B/32). This improvement is particularly pronounced with the CLIP ViT-B/32 backbone, which is trained on a large corpus containing image-text pairs.

\section{Time Complexity Analysis}

\begin{table}[t]
\centering
 \resizebox{0.85\linewidth}{!}{
    \begin{tabular}{@{}lc@{}}
        \toprule
        Variants & Inference Time (ms/per query) \\
        \midrule
        LoRA & 1.16 \\
        Bias Tuning & 1.16 \\
        Prompt Tuning & 1.19 \\
        \midrule
        CLoRA~($N=10$) & 1.17 \\
        CLoRA~($N=1,000$) & 1.18 \\
        CLoRA~($N=100,000$) & 1.26 \\
        \bottomrule
    \end{tabular}
     }
    \caption{Inference time~(ms) per image comparison of PEFT techniques.}
    \label{tab:inference_time_peft}
\end{table}

To delve deeper into the time complexity of KALAHash for highlighting the efficiency and scalability of our approach, we provide the inference time compared to LoRA~\cite{ICLR_LORA}, Bias Tuning~\cite{ACL_BITFIT}, and Prompt Tuning~\cite{IJCV_COOP}. From Table~\ref{tab:inference_time_peft}, LoRA and Bias Tuning do not alter the architecture or inputs. They do not introduce any overhead. Prompt Tuning adds learnable tokens, resulting in a slight increase in computational time. Our KALAHash does introduce some overhead in Equation~\eqref{eq:select} and Equation~\eqref{eq:weight_adjustment}, but these operations are simple and add negligible inference time~($0.01-0.05$ ms) across various architectures. We compare the inference times of KALAHash with other PEFT techniques, considering the knowledge pool size $N$, to demonstrate its efficiency.

\section{More Results}
\begin{table}[t]
\centering
\begin{tabular}{lc}
\toprule
Method & mAP@1000 \\ 
\midrule
\textit{ImageNet-100} \\
MDSH~\cite{CVPR_MDSH} & 24.69 \\
KALAHash & \textbf{30.77}\\
\midrule
\textit{CUB-200} \\
ConceptHash~\cite{DBLP:conf/cvpr/NgZS022} & 1.65 \\
KALAHash & \textbf{9.54}\\
\bottomrule
\end{tabular}
\caption{mAP@1000 on ImageNet-100 and CUB-200.}
\label{tab:imagenet_cub}
\end{table}

Following \citet{ICCV_Hashnet}, we report mAP@1000 on the 1-shot ImageNet-100 dataset. We also ompare our method to ConceptHash~\cite{DBLP:conf/cvpr/NgZS022}, the best paper award of CVPRW24 FGVC11, on the 1-shot CUB-200 dataset. The results are shown in Table~\ref{tab:imagenet_cub}.

\end{document}